\documentclass[letterpaper, 10 pt, conference]{ieeeconf}  % Comment this line out if you need a4paper

\IEEEoverridecommandlockouts                              % This command is only needed if 
\usepackage{dsfont}
\usepackage{graphicx}
\usepackage{amsmath} % Needs to be defined before amsthm package
\usepackage{amssymb}
\usepackage{bm}

\usepackage[table]{xcolor}

\newtheorem{thm}{Theorem}

\newtheorem{defn}{Definition}

\newtheorem{ass}{Assumption}

\newcommand{\R}{{\mathbb R}}

\newcommand{\Rset}{\mathbb{R}}
\newcommand{\Sset}{\mathbb{S}}

\newcommand{\Acal}{{\cal A}}

\newcommand{\Mcal}{{\cal M}}

\newcommand{\Pcal}{{\cal P}}

\newcounter{l1}
\newcounter{l2}
\newcounter{l3}
\newcommand{\bdotlist}{\begin{list}{$\bullet$}{}}
\newcommand{\bboxlist}{\begin{list}{$\Box$}{}}
\newcommand{\bbboxlist}{\begin{list}{\raisebox{.005in}{{\tiny
$\blacksquare$ \ \ }}}{}}
\newcommand{\bdashlist}{\begin{list}{$-$}{} }
\newcommand{\blist}{\begin{list}{}{} }
\newcommand{\barablist}{\begin{list}{\arabic{l1}}{\usecounter{l1}}}
\newcommand{\balphlist}{\begin{list}{(\alph{l2})}{\usecounter{l2}}}
\newcommand{\bAlphlist}{\begin{list}{\Alph{l2}.}{\usecounter{l2}}}
\newcommand{\bdiamlist}{\begin{list}{$\diamond$}{}}
\newcommand{\bromalist}{\begin{list}{(\roman{l3})}{\usecounter{l3}}}

\newcommand{\Eset}{\mathds{E}}
\newcommand{\cvar}{\textrm{CVaR}}
\newcommand{\PNom}{P_0}
\newcommand{\borel}{\Mcal}
\newcommand{\lmax}{\bar{\lambda}}

\title{\LARGE \bf 
Wasserstein Distributionally Robust Risk-Sensitive Estimation via Conditional Value-at-Risk
}

\author{Feras Al Taha$^{1}$ and Eilyan Bitar$^{1}$% <-this % stops a space
\thanks{
This work was supported in part by the Cornell Atkinson Center for Sustainability and in part by the Bezos Earth Fund.
}% <-this % stops a space
\thanks{$^{1}$Feras Al Taha and Eilyan Bitar are with the School of Electrical and Computer Engineering,
        Cornell University, Ithaca, NY 14853 USA 
        {\tt\small foa6@cornell.edu, eyb5@cornell.edu}}%
}

\begin{document}

\maketitle
\thispagestyle{empty}
\pagestyle{empty}

\makeatletter
\setlength{\@fptop}{0pt} % Removes the excessive top glue on float-pages
\makeatother

%%%%%%%%%%%%%%%%%%%%%%%%%%%%%%%%%%%%%%%%%%%%%%%%%%%%%%%%%%%%%%%%%%%%%%%%%%%%%%%%
\begin{abstract}
We propose a distributionally robust approach to risk-sensitive estimation of an unknown signal $\bm{x}$ from an observed signal $\bm{y}$. 
The observation and unknown signal are modeled as random vectors whose joint probability distribution is unknown, but assumed to belong to a given type-2 Wasserstein  ball of distributions, termed the ambiguity set.
The performance of an estimator is measured according to the conditional value-at-risk (CVaR) of the squared estimation error. Within this framework, we study the problem of computing affine estimators that minimize the worst-case CVaR over all distributions in the given ambiguity set. As our main result, we show that, when the  nominal distribution at the center of the Wasserstein ball is finitely supported, such estimators can be exactly computed by solving a tractable semidefinite program. We evaluate the proposed estimators on a wholesale electricity price forecasting task using real market data and show that they deliver lower  out-of-sample CVaR of squared error compared to existing  methods.
\end{abstract}

%%%%%%%%%%%%%%%%%%%%%%%%%%%%%%%%%%%%%%%%%%%%%%%%%%%%%%%%%%%%%%%%%%%%%%%%%%%%%%%%

% *** MAIN FILES ****
\section{Introduction} \label{sec:introduction}

In many decision-making applications that rely on forecasts, guarding against rare but severe prediction errors is more important than minimizing average error alone. In power systems, for example, large errors in load or renewable generation forecasts can increase operating costs and, in extreme cases, threaten system reliability. In financial trading, failing to anticipate sharp price movements can lead to substantial losses or missed profit opportunities. 
Such considerations highlight the need for prediction models that explicitly guard against large  errors, even at the expense of a modest increase in average prediction error.

To better control the upper tail of the prediction error distribution,  one can replace the commonly used mean squared error criterion with a risk-sensitive criterion, such as the \emph{conditional value-at-risk} (CVaR) of the squared error.
However, the effectiveness of such an approach depends on how well the assumed (nominal) distribution, typically estimated from data, captures the events that give rise to large prediction errors. 
If those events are misrepresented in frequency or severity, then minimizing CVaR under the nominal distribution may still leave the predictor exposed to substantial tail risk.
To address this limitation, we incorporate distributional robustness into the risk-sensitive estimation framework by evaluating predictors according to their worst-case CVaR of squared error over a Wasserstein ball centered at the nominal distribution.

\subsubsection*{Related Work}
Robust and distributionally robust approaches to estimation have been studied extensively. For estimation problems in which the latent signal is observed via a linear measurement model with additive noise, robust affine estimators can be obtained by minimizing the worst-case mean squared error (MSE) over uncertainty sets describing the signal or noise, often resulting in tractable semidefinite programming (SDP) reformulations under suitable assumptions \cite{eldar2008minimax, juditsky2018near, beck2007regularization, eldar2004robust, beck2007mean}.  In the distributionally robust setting, one instead minimizes the worst-case MSE over ambiguity sets of probability distributions. Tractable formulations have been derived for ambiguity sets defined by spectral constraints on the signal and noise covariance matrices \cite{eldar2006robust, beck2006robust} and divergence-based constraints on the signal and noise distributions  \cite{levy2004robust, zorzi2017robustness}. Most closely related to our work, \cite{nguyen2023bridging} studies Wasserstein distributionally robust MSE minimization for linear measurement models and establishes optimality of affine estimators (among all possibly nonlinear estimators) when the nominal distribution is elliptical. Wasserstein-based distributionally robust optimization has also been applied in related settings such as Kalman filtering, linear regression, classification, and domain adaptation \cite{shafieezadeh2018wasserstein, shafieezadeh2015distributionally, shafieezadeh2019regularization, aolaritei2026wasserstein, taskesen2021sequential}. 

CVaR-based criteria have been widely used to incorporate risk sensitivity in machine learning and decision-making applications, including multi-armed bandits, reinforcement learning, robust control, fairness-aware learning, and supervised learning \cite{sani2012risk, chow2015risk, chow2018risk, williamson2019fairness, soma2020statistical,laguel2021superquantiles,  van2015distributionally},  to name a few applications. Distributionally robust CVaR minimization with quadratic objectives has also been studied in the context of stochastic control and portfolio optimization problems under moment-based or Gelbrich ambiguity sets \cite{chapman2021toward, kishida2022risk,nguyen2021mean}. 
However, to the best of our knowledge, the problem of Wasserstein distributionally robust CVaR minimization with quadratic loss has not been studied previously.

\subsubsection*{Summary of Contributions}
We study the design of affine risk-sensitive estimators via distributionally robust optimization. 
Specifically, we formulate the estimation problem as the minimization of the worst-case CVaR of the squared estimation error over a type-2 Wasserstein ambiguity set centered at a given nominal distribution. Using duality  for optimal transport problems, we derive an equivalent convex reformulation for the class of worst-case CVaR problems involving generic quadratic loss functions.  This yields an exact reformulation of the estimation problem as a tractable semidefinite program when the nominal distribution is finitely supported. Numerical experiments on an electricity price forecasting task show that the proposed estimators exhibit improved out-of-sample CVaR of squared error relative to existing  methods.

\subsubsection*{Organization}
The remainder of the paper is organized as follows.
Section \ref{sec:formulation} formulates the distributionally robust risk-sensitive estimation problem.
In Section \ref{sec:reformulation}, we derive an equivalent reformulation for this problem as a semidefinite program.
Numerical experiments are provided in Section \ref{sec:experiments}.
We conclude the paper in Section \ref{sec:conclusion}.
All proofs are provided in the appendix.

\subsubsection*{Notation}
Let $\Rset$ and $\Rset_+$ denote the set of real numbers and nonnegative real numbers, respectively.
Let $\Sset^n$ denote the set of all symmetric matrices in $\Rset^{n \times n}$. 
Denote the cone of $n \times n$  real symmetric positive definite (resp. semidefinite) matrices by $\Sset_{++}^{n}$ (resp. $\Sset_+^{n}$). 
Given matrices $A, B \in \Sset^n$, the relation $A \succ B$ (resp. $A\succeq B$) means $A - B \in \Sset^n_{++}$ (resp. $A-B\in\Sset_+^n$). 
Let $I_n$ denote the $n\times n$ identity matrix.
Given a scalar $x\in\Rset$, we let  $(x)_+ := \max\{0,x\}$ represent the positive part of $x$.
Let $\|\cdot\|$ denote the Euclidean norm.
Let $\borel(\Rset^n)$ be the collection of Borel probability measures on $\Rset^n$ with finite second moments. 
We use boldface symbols to denote random variables, and non-boldface symbols to denote particular values in the range of a random variable and other deterministic quantities.

\section{Problem Formulation}  \label{sec:formulation}
We now formulate the distributionally robust risk-sensitive estimation problem studied in this paper.
To begin, suppose that  $\bm{x}$ and $\bm{y}$ are random vectors taking values in $\mathbb{R}^n$ and $\mathbb{R}^m$, respectively, and let $d := n+m$.  
Our objective is to estimate the latent vector $\bm{x}$ from the observation vector $\bm{y}$, where $\bm{y}$ may denote either raw measurements or a preselected set of features extracted from raw data.
The joint  distribution $P \in \Mcal(\Rset^{d})$  of $\bm{z} := (\bm{x},\,\bm{y})$ is unknown, but belongs to a prescribed family of distributions $\Pcal \subseteq \Mcal(\Rset^{d})$, which we refer to as the \emph{ambiguity set}. In this paper, the ambiguity set is specified in terms of the type-2 Wasserstein distance, which is defined as follows.
\begin{defn}
    The \emph{type-2 Wasserstein distance} between two distributions $P_1, P_2 \in \borel(\Rset^d)$  is defined as 
    \begin{align*}
        W_2(P_1, \, P_2)^2 := \hspace{-.075in}\inf_{\pi \in \Pi(P_1, \, P_2)}  \int_{\Rset^d \times \Rset^d} \hspace{-.05in} \| z_1 - z_2 \|^2 \pi(d z_1, \, d z_2),
    \end{align*}
     where $\Pi(P_1, \, P_2)$ denotes the set of all joint distributions  in $\borel(\Rset^d \times \Rset^d)$ with marginal distributions $P_1$ and $P_2$.
\end{defn}

Given a \emph{nominal distribution} $\PNom \in \borel( \Rset^d)$, we define the ambiguity set $\Pcal$ as the set of all distributions whose type-2 Wasserstein distance to $\PNom$ is at most $r \geq 0$, i.e., 
\begin{align} \label{eq:amb_set}
    \Pcal := \big\{ P \in \borel( \Rset^d) \,\big | \, W_2(P, \PNom) \leq r \big\}.
\end{align}
The \emph{radius of the ambiguity set} $r$ represents  the confidence one has in the accuracy of the nominal distribution. In particular, the ambiguity set becomes a singleton containing only the nominal distribution 
when the radius is set to  zero.

Having specified the distributional ambiguity model, we now introduce the class of estimators under consideration. 
We restrict our attention to \emph{affine estimators} of the form 
\begin{align}
    \psi(\bm{y}) = A\bm{y} +b,
\end{align}
where $A \in \Rset^{n \times m}$ and  $b \in \Rset^n$. 
Although affine in  $\bm{y}$, the estimator may depend nonlinearly on the raw data through the feature map defining $\bm{y}$.
The set $\Acal$ denotes the family of all affine mappings from $\Rset^m$ to $\Rset^n$.
To evaluate the quality of an estimator $\psi \in \Acal$, we measure the discrepancy between the estimate $\psi(\bm{y})$ and the target $\bm{x}$ via the \emph{squared error loss}:
\begin{align}
 \ell_\psi(\bm{z}) :=  \|\bm{x} - \psi(\bm{y}) \|^2.   
\end{align}

Rather than minimizing the \emph{mean squared error} (MSE), however, we adopt a risk-sensitive criterion based on the \emph{conditional value-at-risk} (CVaR), which measures the average loss in the upper tail of the loss distribution and therefore places greater emphasis on large estimation errors.
The CVaR is formally defined as follows.
\begin{defn}  \rm Given a random vector $\bm{z} \sim P \in \Mcal(\Rset^d)$ and a Borel-measurable loss function $\ell:\Rset^d \rightarrow \Rset$, the \emph{conditional value-at-risk} (CVaR) of  the loss $\ell(\bm{z})$  at a   probability level $\alpha \in (0,1]$ is defined as:
    \begin{align*}
        \text{CVaR}_P^\alpha \big( \ell(\bm{z}) \big) := \inf_{\tau\in\Rset} \, \Big\{ \tau \, + \, \frac{1}{\alpha} \Eset_P\big[ \big(\ell (\bm{z})-\tau\big)_+ \, \big] \Big\}.
    \end{align*}
\end{defn}

In this paper, we aim to find an affine estimator that minimizes the worst-case CVaR of the squared error loss over all probability distributions in the ambiguity set $\Pcal$, namely,
\begin{align} \label{eq:drrae}
    \inf_{\psi \in\Acal} \sup_{P\in\Pcal} \cvar_P^\alpha\big(  \ell_\psi(\bm{z})  \big).
\end{align}

This formulation subsumes several estimation frameworks as special cases.  For example, when $\alpha = 1$, the CVaR of the squared error coincides with its expectation, and problem \eqref{eq:drrae} reduces to a distributionally robust mean squared error minimization problem of the kind studied in \cite{nguyen2023bridging,shafieezadeh2018wasserstein}. 
At the other extreme, as $\alpha \downarrow 0$, the CVaR converges to the essential supremum of the loss, so  smaller values of $\alpha$ place greater emphasis on extreme estimation errors.
Finally, when the ambiguity set radius is zero, problem \eqref{eq:drrae} reduces to an ``ambiguity-free'' CVaR minimization problem under the nominal distribution $P_0$.

\section{Convex Reformulation} \label{sec:reformulation}
In this section, we derive an equivalent dual reformulation of the worst-case CVaR problem (the inner maximization) in \eqref{eq:drrae}. Using this dual reformulation, we go on to show that, when the nominal distribution is finitely supported, problem \eqref{eq:drrae} can be equivalently reformulated as a tractable semidefinite program (SDP).

\subsection{Dual Reformulation of Worst-Case CVaR} \label{sec:strong_dual}

To solve problem \eqref{eq:drrae}, we first appeal to a strong duality result for generic worst-case risk problems over Wasserstein-based ambiguity sets. This supporting result, adapted from \cite[Theorem 5.22]{Kuhn_Shafiee_Wiesemann_2025}, gives conditions on the loss function under which strong duality is guaranteed to hold for the corresponding worst-case risk problem, and provides a characterization of the dual problem as a two-dimensional convex optimization problem. We then specialize this result to the class of quadratic loss functions considered in this paper. 
The required regularity condition is stated next.\\[0.01cm]
\begin{ass} \label{ass:obj}
    The loss function $\ell:\Rset^d\to\Rset$ is  upper semicontinuous and satisfies  $\sup_{P\in\Pcal} \Eset_P[(\ell(\bm{z}))_+] < \infty$  and $\Eset_P[\ell(\bm{z})]>-\infty$ for all $P\in\Pcal$ and $r > 0$.
\end{ass}

\begin{thm}[Adapted from {\cite[Theorem 5.22]{Kuhn_Shafiee_Wiesemann_2025}}] \label{thm:cvar_dual}  
Let $r>0$ and $\alpha\in(0,\,1]$. If the loss function $\ell:\Rset^d\to\Rset$ satisfies Assumption \ref{ass:obj}, then 
        \begin{align} \label{eq:strong_duality}
           \sup_{P \in \Pcal} \cvar_P^\alpha \big ( \ell(\bm{z}) \big )  
           =  \inf_{\substack{\tau\in\Rset\\ \gamma \in \Rset_+}} \! \tau  +  \frac1\alpha \big( \gamma r^2  + \Eset_{\PNom}\left[  \phi(\tau,\,\gamma, \, \bm{z}) \right] \big) ,
        \end{align}
        where $\phi: \Rset \times \Rset \times \R^d \rightarrow \Rset \cup \{+\infty\}$ is defined as
        \begin{align} \label{eq:zeta} 
          \phi(\tau,\, \gamma, \, \bm{z})  := \sup_{v \in \Rset^d}  \left \{ \big( \ell(v) - \tau\big)_+  \, -  \, \gamma \|  v - \bm{z}  \|^2  \right \}.
        \end{align}
    \end{thm}

Building on Theorem \ref{thm:cvar_dual}, we now provide an equivalent reformulation of worst-case CVaR problems involving (possibly nonconvex) quadratic loss functions.
\begin{thm} \label{thm:cvar_quad_dual}
    Let $r>0$ and $\alpha\in(0,\,1]$, and consider a family of quadratic worst-case CVaR problems given by 
    \begin{align} \label{eq:cvar_quad}
      \sup_{P \in \Pcal} \cvar_P^\alpha \big ( \bm{z}^\top Q \bm{z} + 2q^\top \bm{z} \big),
    \end{align}
    where $Q\in\Sset^d$ and $q\in\Rset^d$. 
    The optimal value of \eqref{eq:cvar_quad} is finite and equal to the optimal value of the following convex program: 
    \begin{align}
        \inf_{\gamma \in\Gamma} \ &  \cvar_{\PNom}^\alpha  \Big (  (\gamma \bm{z} + q)^\top Q_\gamma^{-1}(\gamma \bm{z} + q)  \, +  \,  \gamma \Big( \frac{r^2}{\alpha}  -  \|\bm{z}\|^2 \Big) \Big ), 
        \label{eq:cvar_quad_dual} 
    \end{align}
    where $\Gamma:= \{\gamma \in\Rset_+ \,|\, Q_\gamma \succ 0\}$ and $Q_\gamma := \gamma I_d - Q$. 
\end{thm}

When  $\alpha = 1$, the worst-case CVaR  of the quadratic loss function in \eqref{eq:cvar_quad} reduces to its worst-case expectation.
In this risk-neutral special case, the dual reformulation provided in Theorem \ref{thm:cvar_quad_dual} coincides with previously derived dual reformulations for quadratic worst-case expectation problems under type-2 Wasserstein ambiguity sets \cite{kuhn2019wasserstein,altaha2023distributionally}.
A key difference, however, is that Theorem \ref{thm:cvar_quad_dual} establishes this dual equivalence under weaker assumptions on the nominal distribution. In particular, existing results require the nominal distribution to be either elliptical \cite[Theorem 16]{kuhn2019wasserstein} or absolutely continuous with respect to the Lebesgue measure \cite[Theorem 2]{altaha2023distributionally}, whereas Theorem 2 holds for any nominal distribution $\PNom$ with finite second moments. 

For $\alpha\in(0,\,1)$,  Nguyen et al. \cite{nguyen2021mean} study  a related class of quadratic worst-case CVaR problems using ambiguity sets defined by the Gelbrich distance, which measures the discrepancy between  distributions only through their  means and covariance matrices.\footnote{As shown in \cite{gelbrich1990formula}, the Gelbrich distance between two distributions is a lower bound on their type-2 Wasserstein distance, with equality holding when the distributions being compared are multivariate Gaussian.}
Consequently, the dual reformulation of the quadratic worst-case CVaR problem under Gelbrich ambiguity provided in \cite[Theorems 9 and 10]{nguyen2021mean} depends on the nominal distribution only through its first and second moments.
In contrast,  under type-2 Wasserstein ambiguity,  the dual reformulation  provided in Theorem \ref{thm:cvar_quad_dual}  may depend on higher-order moments of the nominal distribution, beyond its mean and covariance.

\subsection{Semidefinite Programming Reformulation}

Building on Theorem \ref{thm:cvar_quad_dual}, we now show that problem \eqref{eq:drrae} admits an equivalent SDP reformulation when the nominal distribution is finitely supported.

\begin{thm}[SDP Reformulation] \label{thm:sdp}
    Let $r>0$  and $\alpha \in (0, 1]$. If the  nominal distribution $\PNom\in\Mcal(\Rset^d)$ is the uniform distribution on the set $\{z_1,\dots,z_N\}\subseteq\Rset^d$, then problem \eqref{eq:drrae} can be equivalently reformulated as the following SDP:
    \begin{subequations} \label{eq:sdp}
    \begin{align}
        \inf~  & \ \tau+ \frac1\alpha \Big( \gamma r^2 + \frac{1}{N} \sum_{i=1}^N s_i \Big) \label{eq:sdp_obj} \\
        \text{ s.t. }
       & \ A\in\Rset^{n\times m},\, b\in\Rset^n,\, \gamma\in \Rset_+,\, \tau\in\Rset,\, s\in\Rset^N_+,
       \nonumber \\
        & \hspace{0.25mm}\begin{bmatrix}
            \gamma I_d & F^\top\\
            F & I_n
        \end{bmatrix}\succ 0, \label{eq:lmi 1}\\
       & \begin{bmatrix}
            (\tau + s_i\!+\gamma\|z_i\|^2) & \gamma z_i^\top & -b^\top\\
            \gamma z_i & \gamma I_d & F^\top \\
            -b & F & I_n
        \end{bmatrix}\succeq 0,  \forall i=1,\dots,N,\label{eq:lmi 2}
    \end{align}
    \end{subequations}
    where $F := [-I_n,\, A]$, and the decision variables are $A$, $b$, $\gamma$, $\tau$, and $s$.
\end{thm}

\section{Example} \label{sec:experiments}

We consider an electricity price forecasting task where the goal is to predict the vector of hourly day-ahead (DA) energy prices $\bm{x}\in\Rset^{24}$ (\$/MWh) from a feature vector $\bm{y}\in\Rset^{48}$, which contains the corresponding vector of hourly DA load forecasts (MW) provided by PJM, and their squared values.
This setting provides a natural testbed for evaluating risk-sensitive estimators, since electricity prices can exhibit sharp spikes that may not be well represented in historical data.

\subsubsection*{Electricity Market Data} 
\begin{figure}
    \centering
    \phantom{.}\\[0.1cm]
    \includegraphics[width=0.925\linewidth,,trim=1.8cm 0.0cm 2.5cm 0.3cm,clip]{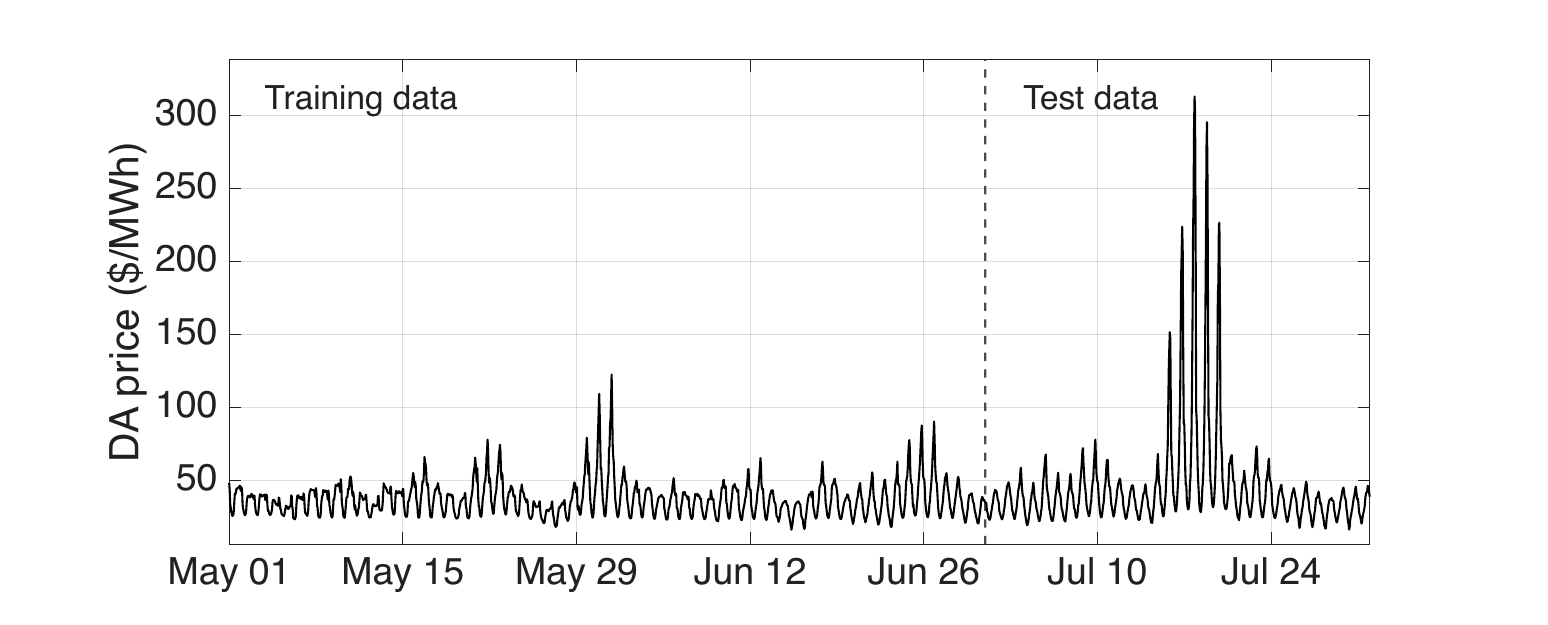}\\
    \includegraphics[width=0.925\linewidth,,trim=1.8cm 0.0cm 2.5cm 0.3cm,clip]{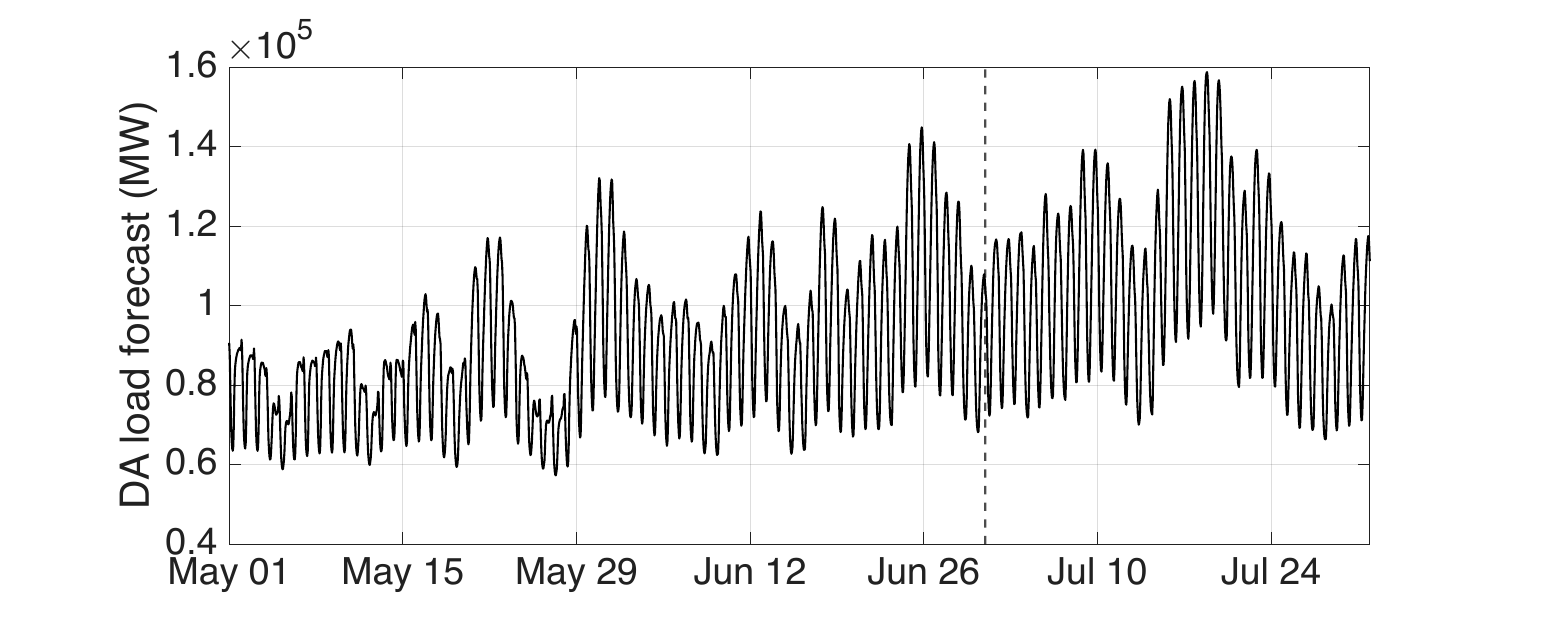}
    \caption{PJM's hourly DA energy prices (top) and DA load forecasts (bottom) from May 1, 2013 to July 31, 2013. The dataset is split between training (before July 1, 2013) and testing (from July 1, 2013 onward) data.}
    \label{fig:data}
\end{figure}

We use the same PJM electricity market dataset as in \cite{donti2017task}, spanning from May 1, 2013 to July 31, 2013. The first two months serve as the training set, and the final month is used for testing. We intentionally choose this test period to include price spikes that exceed those seen in the training data (as shown in Fig.~\ref{fig:data}), thereby providing a demanding out-of-sample assessment of tail-risk performance.
As a preprocessing step, both prices and load forecasts are  rescaled to the unit interval using min-max normalization
computed from the training data, and the same normalization is applied to the test data.
The nominal distribution $\PNom$ is then taken to be the empirical distribution of the normalized training data.

\subsubsection*{Experiments and Discussion}
We compute the proposed distributionally robust CVaR-based estimator (DR-CVaR) for a risk level $\alpha = 0.01$ by solving problem \eqref{eq:sdp}, and compare its out-of-sample performance on the test set with that of the distributionally robust MSE estimator (DR-MSE), obtained by solving the same problem \eqref{eq:sdp} with $\alpha = 1$.
To examine the effect of the ambiguity set radius $r$ on out-of-sample performance, we vary $r$ from $10^{-5}$ to $10^5$, recomputing both estimators for each radius value.

\begin{figure}
    \centering
    \phantom{.}\\[0.1cm]
    \includegraphics[width=0.925\linewidth,,trim=1.8cm 0.0cm 2.5cm 0.3cm,clip]{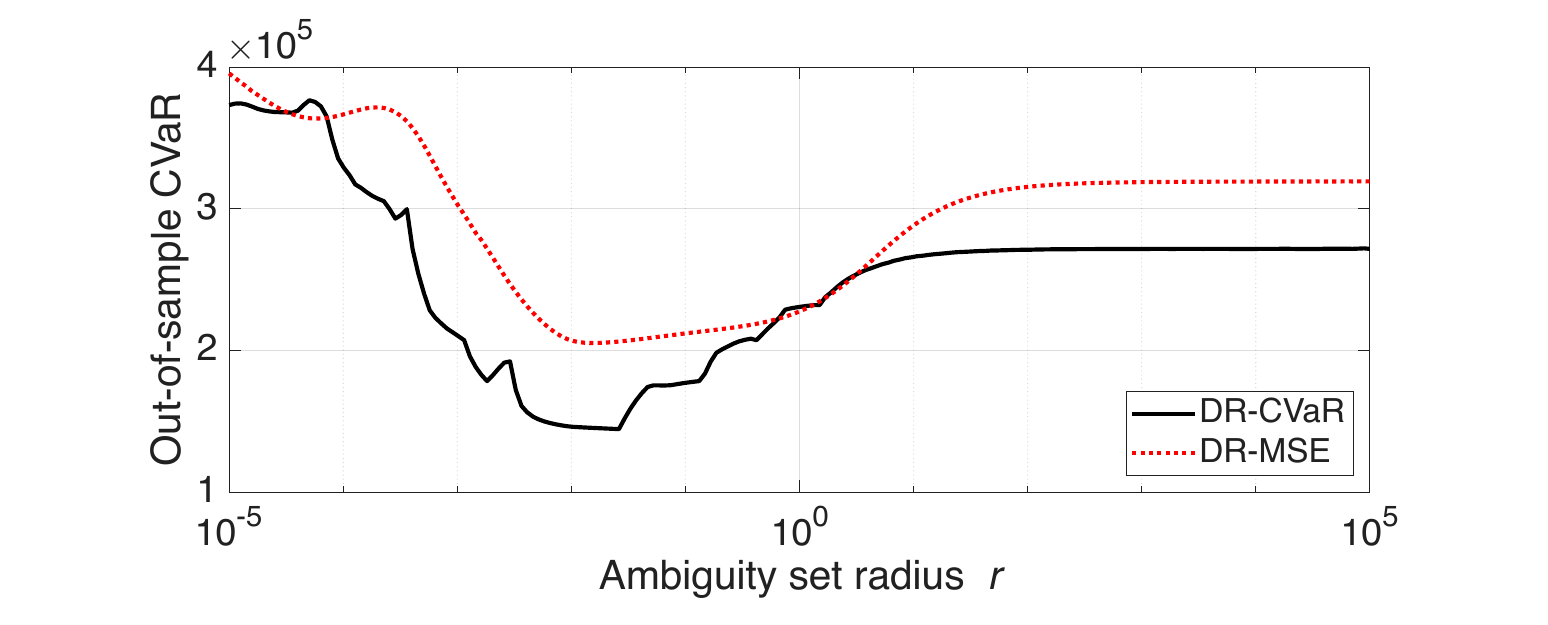}\\
    \includegraphics[width=0.925\linewidth,,trim=1.8cm 0.0cm 2.5cm 0.3cm,clip]{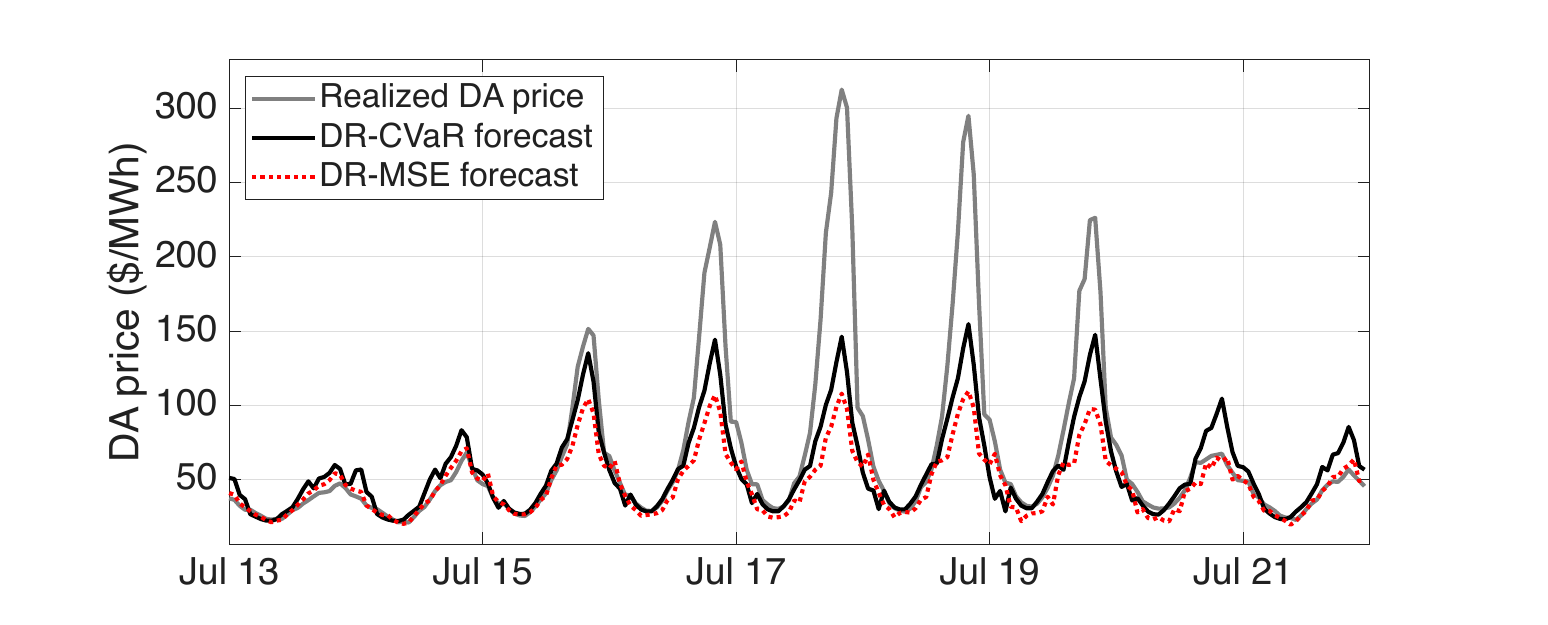}\\
    \caption{
    Top: Out-of-sample CVaR  ($\alpha = 0.01$) of squared error versus the ambiguity set radius $r$ for each estimator,  evaluated on the test data.
    Bottom: Realized DA energy prices (solid gray); DR-CVaR price forecasts (solid black) for $r=0.026$ and $\alpha=0.01$; and DR-MSE price forecasts (dotted red) for $r=0.015$, for July 13--22, 2013.
    The radius values in the bottom panel are chosen to minimize the out-of-sample CVaR depicted in the top panel for illustrative purposes.
    }
    \label{fig:results}
\end{figure}

Fig. \ref{fig:results} (top) depicts the out-of-sample performance of the two estimators, measured by the CVaR of the squared error at $\alpha=0.01$, as a function of the ambiguity set radius.
As $r\downarrow 0$, both estimators reduce to their nominal, non-robust counterparts.
As $r$ initially increases, both estimators benefit from distributional robustness and improve upon their nominal counterparts in terms of out-of-sample CVaR. 
Since the ground-truth distribution is unknown in practice, the radius that minimizes the population CVaR cannot be computed directly, but it can be selected using cross-validation methods.
Across the entire range of radius values considered here, the DR-CVaR estimator consistently achieves a lower out-of-sample CVaR than the DR-MSE estimator.

This behavior is also visible in the forecasts themselves. As shown in Fig. 2 (bottom), the DR-CVaR estimator produces more accurate DA price forecasts on days with larger  price spikes.
This reduction in tail risk comes with a modest loss in  accuracy on days with more typical price patterns, reflecting the intended tradeoff of the risk-sensitive formulation.

\section{Conclusion} \label{sec:conclusion}

In this paper, we proposed a framework for distributionally robust risk-sensitive affine estimation in which the joint distribution of the observation and latent signal is unknown, but assumed to belong to a type-2 Wasserstein ambiguity set. 
Estimators are designed to minimize the worst-case CVaR of the squared estimation error over all distributions in the ambiguity set. By exploiting strong duality in Wasserstein distributionally robust optimization, we derived an equivalent convex reformulation of the resulting worst-case CVaR problem. 
This leads to an equivalent reformulation of the proposed estimation problem as a tractable semidefinite program when the nominal distribution is finitely supported.
A natural direction for future research is to extend the proposed framework to nonlinear estimator classes, such as neural networks.

\appendix

\section{Deferred Proofs}

\subsection{Proof of Theorem \ref{thm:cvar_quad_dual}}

The proof proceeds in three steps. First, we show that strong duality, as stated in Theorem \ref{thm:cvar_dual}, holds for the class of quadratic worst-case CVaR problems specified in \eqref{eq:cvar_quad}. Second, we show that both the primal and dual optimal values are finite. Third, we prove the equivalence between problems \eqref{eq:cvar_quad} and \eqref{eq:cvar_quad_dual}. Throughout the proof, we denote the quadratic loss function by $\ell(z):=z^\top Qz+2q^\top z$. 

\

%%%%%%%%%%%%%%%%%%%%%%%%%%%%%%%%%%%%%%%
 \emph{Step 1 (Establishing strong duality):} By Theorem  \ref{thm:cvar_dual}, strong duality holds if the given loss function satisfies Assumption~\ref{ass:obj}. Since the loss function $\ell$ is quadratic, there exists a finite constant $C>0$ such that $|\ell(z)| \le C\bigl(1+\| z\|^2\bigr)$ for all $ z\in\Rset^d.$ Moreover, for every $P\in\Pcal$, it holds that $\Eset_P[\|\bm z\|^2]  \leq  2r^2 + 2\Eset_{\PNom}[\|\bm z\|^2]$. Hence, for every $P\in\Pcal$, we have the uniform upper bound:
\[
\Eset_P[|\ell(\bm z)|]
\le
C\bigl(1+\Eset_P[\|\bm z\|^2]\bigr)
\le
C\bigl(1+2r^2+2\Eset_{\PNom}[\|\bm z\|^2]\bigr).
\]
Since $\Eset_{\PNom}[\|\bm z\|^2] < \infty$ by assumption, the above inequalities imply
$\sup_{P\in\Pcal}\Eset_P[|\ell(\bm z)|] < \infty$, verifying Assumption $\ref{ass:obj}$. 

\

%%%%%%%%%%%%%%%%%%%%%%%%%%%%%%%%%%%%%%%
\emph{Step 2 (Finiteness of primal and dual optimal values):}
We prove finiteness of the primal and dual optimal values by showing that they are bounded from above and below.
First, the primal optimal value is bounded from below since
\begin{align} \label{eq:lower_bd}
    \sup_{P\in\Pcal} \cvar_P^\alpha \big ( \ell(\bm{z})\big)
    &\ge \Eset_{ \PNom}[\ell(\bm{z})] > - \infty.
\end{align}
The first inequality follows from the fact that  $\PNom\in\Pcal$ and that the CVaR of a random variable is no smaller than its mean. The second inequality follows from Assumption~\ref{ass:obj}, which the quadratic loss  $\ell$ was shown to satisfy  in Step 1.

Next, we show that the dual optimal value is bounded from above. By Theorem \ref{thm:cvar_dual}, we have the primal-dual equivalence:
\begin{align} 
    \label{eq:quadratic_dual} &\sup_{P\in\Pcal} \cvar_P^\alpha \big ( \ell(\bm{z}) \big)
     = \inf_{\substack{\tau\in\Rset\\ \gamma\in\Rset_+}} \tau + \frac1\alpha \big (\gamma r^2 + \Eset_{\PNom}[\phi(\tau,\,\gamma,\,\bm{z})]\big), \\[-0.95cm] \nonumber
\end{align}
where
\begin{align} \label{eq:phi}
\phi(\tau,\,\gamma,\,\bm{z}) := \sup_{v\in\Rset^d} \big\{ (\ell(v)-\tau)_+-\gamma\|v-\bm{z}\|^2 \big\}.
\end{align}
For any $(\tau,\,\gamma,\,z)\in\Rset\times\Rset_+\times\Rset^d$, it holds that
\begin{align*}
    &\phi(\tau,\,\gamma,\,z) \\
    &\hspace{0.1cm} 
    = \sup_{v\in\Rset^d} \Big \{ \max\big( \ell(v) - \tau - \gamma \|v-z\|^2,\, - \gamma \|v-z\|^2\big) \Big\} \\
    &\hspace{0.1cm} \overset{(a)}{=}  \max\Big(\sup_{v\in\Rset^d} \ell(v) - \tau - \gamma \|v-z\|^2,\, \sup_{v\in\Rset^d} - \gamma \|v-z\|^2\Big) \\
    &\hspace{0.1cm} \overset{(b)}{=}  \max\Big(\sup_{v\in\Rset^d} - v^\top Q_\gamma v + 2 (\gamma z + q)^\top v  -\gamma\|z\|^2 - \tau  ,\, 0\Big). 
\end{align*}
Equality $(a)$ follows from interchanging the supremum over $v$ with the $\max$ operator. Equality $(b)$ follows from the fact that $-\gamma\|v-z\|^2$ is maximized at $v=z$ since $\gamma \ge 0$. 

Now, note that for any $\gamma \in \Gamma = \{\gamma \in \Rset_+ \, | \, Q_\gamma \succ 0\}$,  the quadratic function $ v \mapsto -v^\top Q_\gamma v +2(\gamma z+q)^\top v$ is strictly concave in $v$. It  therefore has a unique maximizer, which is given by $v^\star= Q_\gamma^{-1} (\gamma z+q)$. Thus, for every $\gamma \in \Gamma$, the function $\phi(\tau,\,\gamma,\,z)$ simplifies to
\begin{align} \label{eq:phi_soln}
    \phi(\tau,\,\gamma,\,z)\! = \!\big( (\gamma z+q)^\top Q_\gamma^{-1}(\gamma z+q) -\gamma\|z\|^2 \! \! - \tau  \big)_+.
\end{align}
We use this fact to upper bound the dual optimal value as follows. Note that for any $\bar{\gamma} \in \Gamma$, it holds that
\begin{align}
    \nonumber &\inf_{\substack{\tau\in\Rset\\ \gamma\in\Rset_+}} \tau + \frac1\alpha \big (\gamma r^2 + \Eset_{\PNom}[\phi(\tau,\,\gamma,\,\bm{z})]\big)\\
    \nonumber &\hspace{0.4cm} \overset{(a)}{\le} \frac1\alpha \big (\bar{\gamma} r^2 + \Eset_{\PNom}[\phi(0,\,\bar{\gamma},\,\bm{z})]\big)\\
    \nonumber &\hspace{0.4cm} \overset{(b)}{=} \frac1\alpha \big (\bar{\gamma} r^2 + \Eset_{\PNom}[((\bar{\gamma} \bm{z}+q)^\top Q_{\bar{\gamma}}^{-1}(\bar{\gamma} \bm{z}+q) -\bar{\gamma}\|\bm{z}\|^2 )_+]\big)\\
    \label{eq:upper_bd} &\hspace{0.4cm} \overset{(c)}{<} \infty.
\end{align}
Inequality $(a)$ follows from taking $\tau=0$ and $\gamma=\bar{\gamma}$. Equality 
$(b)$ follows from the identity in \eqref{eq:phi_soln}, and $(c)$ follows from the assumption that the nominal distribution $\PNom$ has a finite second moment.
Combining  \eqref{eq:lower_bd}, \eqref{eq:quadratic_dual}, and \eqref{eq:upper_bd}, it follows that  the primal and dual optimal values are finite.

\

%%%%%%%%%%%%%%%%%%%%%%%%%%%%%%%%%%%%%%%
\emph{Step 3 (Reformulating the dual problem):} 
To complete the proof, it remains to show that the right-hand side of \eqref{eq:quadratic_dual} reduces to \eqref{eq:cvar_quad_dual}.
To this end, we show that the infimum in \eqref{eq:quadratic_dual} is unchanged if the feasible set for the decision variable $\gamma$ is restricted from $\Rset_+$ to the set
$$ \Gamma = \{\gamma \in \Rset_+ \, | \, Q_\gamma \succ 0 \} =  \{\gamma \in \Rset_+ \, | \, \gamma >  \lmax(Q) \}, $$ 
where $\lmax(Q)$ denotes the largest eigenvalue of the matrix $Q$. 
To show this, it will be convenient to rewrite the infimum in \eqref{eq:quadratic_dual} as

\begin{align} \label{eq:rewrite_inf}
    \inf_{\substack{\tau\in\Rset\\ \gamma\in\Rset_+}}  \Big\{ \tau + \frac{1}{\alpha} f(\tau, \, \gamma) \Big\},
\end{align}
where $f( \tau, \, \gamma) := \gamma r^2 + \Eset_{\PNom}[\phi(\tau,\,\gamma,\,\bm{z})]$. 

First, it is straightforward to see that if $\gamma < \lmax(Q)$ (equivalently $Q_\gamma \not \succeq 0$), then  $\phi(\tau,\,\gamma,\,z)  = \infty$ for all $(\tau, z) \in \Rset \times \Rset^d$.
Hence, $f(\tau, \, \gamma) = \infty$ for any $\tau \in \Rset$ and $\gamma < \lmax(Q)$, so these values of $\gamma$ may be excluded from the feasible set in problem \eqref{eq:rewrite_inf} without changing the  value of the infimum. 

Next, we show that the point $\gamma = \lmax(Q)$ may also be excluded from the feasible set in problem \eqref{eq:rewrite_inf} without changing the  value of the infimum.
We only consider the case where $\lmax(Q) \ge 0$; otherwise, the point $\gamma = \lmax(Q) < 0$ already lies outside the feasible set of problem \eqref{eq:rewrite_inf}.
Let $\bar{\gamma} \in \Gamma$ and for each $\theta \in (0,1)$, define $$\gamma_\theta := \theta \bar{\gamma} + (1- \theta) \lmax(Q).$$  
Fix $\tau \in \Rset$. 
The function $f(\tau, \gamma) =  \gamma r^2 + \Eset_{\PNom}[\phi(\tau,\,\gamma,\,\bm{z})]$ is convex with respect to $\gamma$. 
Indeed, for each $(\tau, \, z) \in \Rset \times \Rset^d$, the map $\gamma \mapsto \phi(\tau, \, \gamma, \, z)$  in \eqref{eq:phi} is the pointwise supremum of affine functions of $\gamma$, and is therefore convex; and taking expectation with respect to $\bm{z}$ preserves convexity. Using the fact that $\gamma_\theta \in \Gamma$ and the convexity of $f(\tau, \cdot)$, we get 
\begin{align*}
\inf_{\gamma \in \Gamma} f(\tau,\gamma) & \leq f(\tau, \, \gamma_\theta)  \leq  \theta f(\tau,\bar{\gamma})+(1-\theta)f(\tau,\,\lmax(Q)).
\end{align*}
Letting $\theta \downarrow 0$ gives $\inf_{\gamma \in \Gamma} f(\tau,\gamma)  \leq f(\tau,\,\lmax(Q))$ since $f(\tau,\bar{\gamma})$ is finite. Thus, the point $\gamma = \lmax(Q)$ may also be excluded from the feasible set in \eqref{eq:rewrite_inf} without changing the value of the infimum.

We have shown that the feasible set for $\gamma$ in \eqref{eq:rewrite_inf} may be restricted from $\Rset_+$ to $\Gamma$ without changing the value of the infimum. It follows that
\begin{align*}
    &\inf_{\substack{\tau\in\Rset\\ \gamma\in\Rset_+}} \tau + \frac1\alpha f(\tau,\,\gamma) = \inf_{\substack{\tau\in\Rset\\ \gamma\in\Gamma}} \ \tau + \frac1\alpha \big (\gamma r^2 + \Eset_{\PNom}[\phi(\tau,\,\gamma,\,\bm{z})]\big)\\
    &\hspace{0cm} = \inf_{\substack{\tau\in\Rset\\ \gamma\in\Gamma}} \ \tau + \frac1\alpha \big (\gamma r^2\\[-0.3cm]
    &\hspace{1.0cm} + \Eset_{\PNom}\big[\big((\gamma \bm{z}+q)^\top Q_\gamma^{-1}(\gamma \bm{z}+q) -\gamma\|\bm{z}\|^2 - \tau\big)_+\big]\big).
\end{align*}
Here, the second equality follows from \eqref{eq:phi_soln}.
Finally, the infimum in the last line above can be shown to be equal to problem \eqref{eq:cvar_quad_dual} by applying the change of variables $\tau = \bar{\tau} - \gamma r^2/\alpha$, and using the definition of CVaR.

\subsection{Proof of Theorem \ref{thm:sdp}}
For an affine estimator $\psi\in\Acal$, the squared error loss can be written as
\begin{align*}
    \ell_\psi(z) = z^\top F^\top F z +2b^\top F z + b^\top b,
\end{align*}
where $F:= [-I_n,\,A]$.\
Taking $Q=F^\top F$ and $q=F^\top b$, and applying Theorem \ref{thm:cvar_quad_dual}, we can reformulate the worst-case CVaR of the squared error loss as follows
\begin{align*}
    &\sup_{P\in\Pcal} \cvar_P^\alpha(\bm{z}^\top F^\top F\bm{z} + 2b^\top F\bm{z}) +b^\top b\\ 
    &= \!\! \inf_{\gamma\in\Rset_+} \! \! \Big\{ \cvar_{\PNom}^\alpha \! \Big(\!(\gamma \bm{z}+F^\top b)^\top \! (\gamma I_d - F^\top F)^{-1}  (\gamma \bm{z}+F^\top b) \\
    &\hspace{2.15cm}+ \gamma(\alpha^{-1}r^2-\|\bm{z}\|^2) \Big) +b^\top b \, \Big | \, \gamma I_d \succ F^\top F \Big \}.
\end{align*}
Using the assumption that $P_0$ is the uniform distribution on the set $\{z_1,\,\dots,\,z_N\}$, the inner maximization in problem \eqref{eq:drrae} can be rewritten as the following convex program:\vspace{-0.225cm}
\begin{align*}
    \inf \ & \bar{\tau} + b^\top b + \frac1{\alpha N} \sum_{i=1}^N \max\big( \gamma (\alpha^{-1} r^2 -\|z_i\|^2)-\bar{\tau}\\
    & + (\gamma z_i + F^\top b)^\top (\gamma I_d - F^\top F)^{-1} (\gamma z_i + F^\top b) ,\,0\big) \\
    \text{s.t.} \ & \bar{\tau}\in\Rset,\, \gamma\in\Rset_+,\, \gamma I_d - F^\top F \succ 0.
\end{align*}
Using the change of variables $ \bar{\tau}  =  \tau + \gamma r^2/\alpha - b^\top b$ and putting each term in the summation over $i$ in epigraph form, this convex program can be equivalently reformulated as: \vspace{-0.2cm}
\begin{align*}
    \inf \ & \tau+ \frac1\alpha \Big( \gamma r^2 + \frac1N \sum_{i=1}^N  s_i  \Big)\\
    \text{s.t.} \ & \tau\in\Rset,\, \gamma\in\Rset_+,\, s\in\Rset^N, \, \gamma I_d - F^\top F \succ 0,\\
    & s_i \ge \max\big( (\gamma z_i + F^\top b)^\top (\gamma I_d - F^\top F)^{-1} (\gamma z_i + F^\top b) \\ 
    &\qquad  + b^\top b -\gamma\|z_i\|^2-\tau,\,0\big), \quad \forall i=1,\dots,N.
\end{align*}
It follows from the Schur complement condition for positive semidefiniteness that the constraint $\gamma I_d - F^\top F \succ 0$ is equivalent to the linear matrix inequality \eqref{eq:lmi 1}.
Moreover, each constraint involving the variable $s_i$ can be expressed as two constraints: $s_i\ge 0$ and 
\begin{align}
    \nonumber&s_i \ge  b^\top b -\gamma\|z_i\|^2-\tau\\
    \nonumber&\qquad +(\gamma z_i + F^\top b)^\top (\gamma I_d - F^\top F)^{-1} (\gamma z_i + F^\top b) \\
    \label{eq:matrix_ineq} \Longleftrightarrow \ & \begin{bmatrix}
        s_i + \gamma\|z_i\|^2+\tau-b^\top b & (\gamma z_i+F^\top b)^\top\\
        \gamma z_i+F^\top b & \gamma I_d - F^\top F
    \end{bmatrix} \succeq 0,
\end{align}
where the last equivalence follows from using the Schur complement condition for positive semidefiniteness since $\gamma I_d - F^\top F \succ 0$.
To prove equivalence between the matrix inequality \eqref{eq:matrix_ineq} and \eqref{eq:lmi 2}, it is helpful to rewrite \eqref{eq:matrix_ineq} as
\begin{align*}
    \begin{bmatrix}
        s_i + \gamma\|z_i\|^2+\tau & \gamma z_i^\top\\
        \gamma z_i & \gamma I_d 
    \end{bmatrix} - \begin{bmatrix} -b^\top\\ F^\top
    \end{bmatrix} \begin{bmatrix} -b^\top\\ F^\top
    \end{bmatrix}^\top \succeq 0.
\end{align*}
Using the Schur complement condition once again, it can be shown  that the above constraint is equivalent to \eqref{eq:lmi 2}.
This proves that problem \eqref{eq:drrae} is equivalent to the SDP \eqref{eq:sdp}.

%%%%%%%%%%%%%%%%%%%%%%%%%%%%%%%%%%%%%%%%%%%%%%%%%%%%%%%%%%%%%%%%%%%%%%%%%%%%%%%%

% *** BIBLIOGRAPHY ****
\bibliographystyle{IEEEtran}
\bibliography{references}             % bib file to produce the bibliography

\end{document}